%% file: main.tex
\pdfoutput=1

\documentclass[10pt,twocolumn,letterpaper]{article}

\usepackage[pagenumbers]{cvpr}

\input{preamble}

\definecolor{cvprblue}{rgb}{0.21,0.49,0.74}
\usepackage[pagebackref,breaklinks,colorlinks,allcolors=cvprblue]{hyperref}
\hypersetup{pdftitle={SRC-Flow: Compact Semantic Representations Enable Normalizing Flows for Image Generation},pdfauthor={Longtao Jiang, Jianmin Bao, Zhendong Wang, Xin Tao, Zhihui Li, Xiaojun Chang}}

\def\confName{CVPR}
\def\confYear{2027}

\title{SRC-Flow: Compact Semantic Representations Enable\\
Normalizing Flows for Image Generation}

\author{%
\textbf{Longtao Jiang}$^{1}$\hspace{0.45em}
\textbf{Jianmin Bao}$^{2,\dagger}$\hspace{0.45em}
\textbf{Zhendong Wang}$^{1}$\hspace{0.45em}
\textbf{Xin Tao}$^{2}$\hspace{0.45em}
\textbf{Zhihui Li}$^{1}$\hspace{0.45em}
\textbf{Xiaojun Chang}$^{1,\dagger}$\\[0.4em]
{\normalsize $^{1}$Mohamed bin Zayed University of Artificial Intelligence}\qquad 
{\normalsize $^{2}$Kling Team, Kuaishou Technology}
}

\begin{document}

\twocolumn[{%
\renewcommand\twocolumn[1][]{#1}%
\maketitle
\begin{center}
    \centering
    \captionsetup{type=figure}
    \vspace{-0.3em}
    \includegraphics[width=\textwidth,height=0.23\textheight,keepaspectratio]{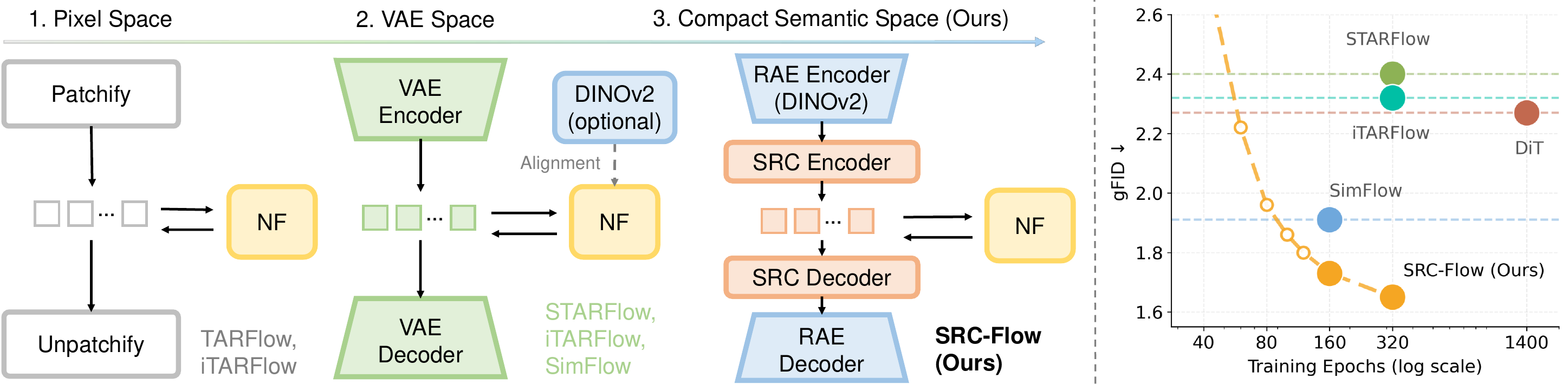}
    \vspace{-1.8em}
    \caption{
    SRC-Flow moves NF modeling from pixels or VAE latents to compact semantic representations produced by the SRC.
    }
    \label{fig:teaser}
\end{center}%
}]

\begingroup\renewcommand{\thefootnote}{}\footnotetext{$^\dagger$Corresponding authors.}\endgroup

\begin{abstract}
Normalizing flows (NFs) provide exact likelihoods and deterministic invertible sampling, but have historically lagged behind diffusion models for large-scale image generation.
We identify a key obstacle: NFs are required to learn a single invertible transport over the full ambient space, making them highly sensitive to high-dimensional representations.
This leads to a semantic-capacity mismatch in modern visual representation spaces, where semantic information is compact but encoded in overcomplete features.
We propose \textbf{SRC-Flow}, which introduces a Semantic Representation Compressor (SRC) to compact high-dimensional RAE features into a low-dimensional semantic space before flow modeling and preserve reconstruction through the frozen RAE decoder.
This compact space reduces the modeling burden of NFs and enables effective likelihood-based generation in semantic representation space.
We further adopt constant noise regularization tailored to the fixed unconditional bijection learned by flows.
On ImageNet $256 \times 256$ and $512 \times 512$, SRC-Flow achieves state-of-the-art generation quality among normalizing flow methods, with gFID scores of 1.65 and 2.07 under classifier-free guidance, while retaining exact likelihood computation in the compact semantic representation space and deterministic invertible sampling at the flow level. Code and models will be released.
\end{abstract}

\section{Introduction}
\label{sec:intro}

Normalizing flows (NFs)~\citep{rezende2015variational, dinh2017density} are generative models with exact likelihoods and deterministic invertible sampling.
This makes NFs theoretically appealing, but their generation quality has long lagged behind diffusion models~\citep{ho2020ddpm, song2021scorebased, song2021ddim, karras2022edm}.
Recent Transformer-based flows have narrowed this gap: TARFlow~\citep{zhai2024tarflow} revisited pixel-space autoregressive flows, STARFlow~\citep{gu2025starflow} scaled NFs to VAE latents, SimFlow~\citep{zhao2025simflow} simplified latent NF training, and iTARFlow~\citep{chen2026itarflow} added iterative denoising.
These advances suggest NFs remain promising but raise a central question: \emph{which representation space is best suited for normalizing flows?}

This choice of representation space is especially important for NFs because their exact likelihood objective requires an invertible transport over the full ambient space.
Unlike diffusion or rectified-flow models, which can redistribute learning across noise levels through time-dependent denoising, an NF learns a single fixed bijection between data and prior.
Thus, every modeled dimension contributes to the likelihood and log-determinant.
Pixel-space NFs suffer from extreme dimensionality, while VAE-latent NFs reduce dimensionality but have limited semantic structure.

Representation Autoencoders (RAEs)~\citep{zheng2025rae} provide an appealing alternative by pairing a frozen pretrained vision encoder, such as DINOv2~\citep{oquab2024dinov2, caron2021dino}, with a trained decoder.
Their representations are semantically rich and structurally organized, but they are also high-dimensional and overcomplete.
This leads to a \emph{semantic-capacity mismatch}: the semantic information needed for generation is compact, whereas an NF must model the full high-dimensional representation through exact invertible transport.
Diffusion models can partially absorb this mismatch through dimension-dependent noise schedule shifts~\citep{esser2024scaling}, but NFs have no timestep or noise schedule mechanism.
Consequently, applying NFs to full RAE representations forces the bijection to model every channel under exact invertibility, including redundant or weakly informative ones, increasing the burden of likelihood training.

We propose \textbf{SRC-Flow}, a normalizing-flow framework built on compact semantic representations.
Instead of directly modeling full RAE tokens, SRC-Flow introduces a \textbf{Semantic Representation Compressor (SRC)} that compresses high-dimensional semantic features from dimension $n$ to a compact dimension $d \ll n$ before flow modeling, while preserving reconstruction through the frozen RAE decoder.
This is motivated by the compact structure of pretrained representations: a few channels preserve most of the semantic variation in RAE features.
By reducing ambient dimensionality while retaining semantic content, SRC lets NFs benefit from pretrained representation spaces without being overwhelmed by their overcomplete structure.

We further find that the per-example noise inherited from RAE decoder training is suboptimal for flows: since an NF learns a single unconditional bijection, modeling a mixture of differently perturbed distributions hinders likelihood training.
We therefore adopt constant noise regularization, which better matches the transport learned by NFs.

Our contributions are threefold.
First, we identify a semantic-capacity mismatch between high-dimensional pretrained representations and exact-likelihood normalizing flows, verified with a naive full-representation baseline.
Second, we propose SRC, which extracts compact semantic representations from frozen RAE features while preserving decoder compatibility.
Third, SRC-Flow achieves state-of-the-art quality among NF methods on ImageNet $256 \times 256$ and $512 \times 512$, with gFID scores of 1.65 and 2.07 under classifier-free guidance, while retaining exact likelihood computation in the compact semantic representation space and deterministic invertible sampling at the flow level.

\section{Related Work}
\label{sec:related}

\paragraph{Normalizing Flows for Image Generation.}
Normalizing flows are exact-likelihood generative models built from tractable invertible transformations.
Coupling flows such as NICE~\citep{dinh2015nice}, RealNVP~\citep{dinh2017density}, and Glow~\citep{kingma2018glow} enabled likelihood modeling with triangular Jacobians, and autoregressive~\citep{papamakarios2017masked, kingma2016improved}, continuous~\citep{chen2018neural, grathwohl2019ffjord}, residual~\citep{behrmann2019residual, chen2019residual}, and spline flows~\citep{durkan2019neural}, as well as Flow++~\citep{ho2019flow++}, further improved expressiveness and training.
Nevertheless, NFs were largely surpassed in image synthesis by diffusion~\citep{ho2020ddpm, song2021scorebased, song2021ddim, karras2022edm} and autoregressive models~\citep{oord2016pixel, esser2021taming, ramesh2021dalle}.
Transformer-based NFs have recently renewed this direction: TARFlow~\citep{zhai2024tarflow} generates images directly in pixel space with masked autoregressive flows, STARFlow~\citep{gu2025starflow} moves NFs to VAE latent space with a deep-shallow architecture, SimFlow~\citep{zhao2025simflow} jointly optimizes the autoencoder and flow, and iTARFlow~\citep{chen2026itarflow} addresses the noise dilemma with multi-noise training and iterative denoising.
Related systems include JetFormer~\citep{tschannen2024jetformer}, FARMER~\citep{zheng2025farmer}, and FlowBack~\citep{chen2025flowing}, which aligns reverse-pass NF features with pretrained visual representations.
In contrast, SRC-Flow focuses on the modeling space itself, placing NFs in a compact semantic space.

\paragraph{Latent Spaces for Generative Models.}
Latent Diffusion Models~\citep{rombach2022ldm} popularized the two-stage VAE-based paradigm~\citep{kingma2014vae}, now widely adopted in high-resolution systems~\citep{podell2024sdxl, ramesh2022dalle2, esser2024scaling}.
Diffusion Transformers~\citep{peebles2023scalable} and interpolant or rectified-flow models~\citep{lipman2023flow, liu2023flow, ma2024sit} further improved scalability, although these ``flow'' objectives generally do not provide exact likelihood through a change-of-variables formula.
Latent generation has also been improved by representation alignment~\citep{yu2024repa}, VAE optimization~\citep{yao2025vavae}, equivariant regularization~\citep{kouzelis2025eqvae}, end-to-end VAE tuning~\citep{leng2025repae}, and decoupled diffusion architectures~\citep{wang2025ddt}.
Representation Autoencoders (RAEs)~\citep{zheng2025rae} replace VAE encoders with pretrained representation models such as DINOv2~\citep{oquab2024dinov2}, SigLIP~\citep{tschannen2025siglip}, and other self-supervised or language-supervised encoders~\citep{dosovitskiy2021vit, caron2021dino, he2022mae, radford2021clip}, whose features are semantically rich but high-dimensional and overcomplete.
AlignTok~\citep{chen2025aligntok}, FlatDINO~\citep{calvogonzalez2026flatdino}, LV-RAE~\citep{liu2026lvrae}, and PS-VAE~\citep{zhang2025psvae} adapt such features for reconstruction or diffusion-based generation, whereas SRC-Flow studies how they should be structured for exact-likelihood NFs.

\section{Preliminaries}
\label{sec:preliminaries}

\paragraph{Representation Autoencoders.}
RAE~\citep{zheng2025rae} pairs a frozen pretrained visual encoder with a trained image decoder.
We use DINOv2-B~\citep{oquab2024dinov2} as the encoder $E$.
Given an image $x \in \mathbb{R}^{3 \times H \times W}$, $E$ produces patch tokens as follows:
\begin{equation}
    z_{\mathrm{raw}} = E(x) \in \mathbb{R}^{N \times n},
    \qquad
    z = \frac{z_{\mathrm{raw}}-\mu_{rae}}{\sigma_{rae}},
    \label{eq:rae_normalize}
\end{equation}
where $N=\frac{H}{p}\times\frac{W}{p}$ is the number of spatial tokens, $n$ is the channel dimension, and $(\mu_{rae},\sigma_{rae})$ are precomputed channel-wise statistics.
The normalized feature $z \in \mathbb{R}^{N \times n}$ is the semantic representation used in this work.
A frozen RAE decoder $D$ reconstructs images from denormalized features.
Compared with VAE latents, RAE features inherit richer semantics from pretrained models, but their high ambient dimension challenges exact-likelihood NFs.

\paragraph{Normalizing flows.}
A normalizing flow~\citep{rezende2015variational,dinh2017density} defines an invertible mapping $f_\theta:\mathcal{Y}\rightarrow\mathcal{U}$ from a data distribution $p_{\mathcal{Y}}$ to a simple prior $p_{\mathcal{U}}=\mathcal{N}(0,I)$, typically composed as $f_\theta=f_{K-1}\circ f_{K-2}\circ\cdots\circ f_0$.
For a modeling variable $y\in\mathcal{Y}$ and its corresponding prior variable $u=f_\theta(y)$, the exact log-likelihood is given by the change-of-variables formula:
\begin{equation}
    \log p_{\mathcal{Y}}(y)
    =
    \log p_{\mathcal{U}}(f_\theta(y))
    +
    \log \left|
    \det \frac{\partial f_\theta}{\partial y}
    \right|.
    \label{eq:nf_objective}
\end{equation}
Generation applies $y=f_\theta^{-1}(u)$ to $u\sim\mathcal{N}(0,I)$.
Unlike diffusion models, which learn time-dependent denoising, NFs learn a single deterministic bijection and optimize exact likelihood over the full modeled space.

\paragraph{Transformer autoregressive flow.}
We instantiate $f_\theta$ with the Transformer Autoregressive Flow (TAF) used in recent flow-based image generation methods~\citep{zhai2024tarflow,gu2025starflow,zhao2025simflow}.
TAF consists of $K$ autoregressive affine blocks.
For the $i$-th token in block $k$, a causal Transformer predicts shift and log-scale parameters $(\mu_i^k,\alpha_i^k)$ from previous tokens and an optional class label.
The forward data-to-prior mapping and reverse sampling step are as follows:
\begin{equation}
    \begin{aligned}
    y_i^{k+1} &= (y_i^k-\mu_i^k)\odot \exp(-\alpha_i^k), \\
    y_i^k &= y_i^{k+1}\odot \exp(\alpha_i^k)+\mu_i^k .
    \end{aligned}
    \label{eq:taf_affine}
\end{equation}
The autoregressive structure yields a tractable triangular Jacobian, and generation proceeds sequentially in the inverse direction.
Following SimFlow~\citep{zhao2025simflow}, we use a \emph{deep-shallow} design with $K=6$ blocks: the first $K-1$ blocks are shallow, while the last block is deep, and enable classifier-free guidance by randomly dropping class labels during training.
In SRC-Flow, the variable $y$ is the compact semantic representation produced by the Semantic Representation Compressor.

\section{Method}
\label{sec:method}

We present \textbf{SRC-Flow}, a framework that enables normalizing flows to model compact semantic representation spaces.
SRC-Flow first compresses high-dimensional pretrained visual features into a compact representation, and then trains an exact-likelihood normalizing flow in this space.
We begin by identifying the semantic-capacity mismatch between representation spaces and normalizing flows (Section~\ref{sec:challenges}), then introduce the Semantic Representation Compressor (SRC) (Section~\ref{sec:src}), the noise regularization strategy (Section~\ref{sec:noise}), and the full training and inference pipeline (Section~\ref{sec:pipeline}).

\subsection{Semantic-Capacity Mismatch in NFs}
\label{sec:challenges}

Modern pretrained visual representations provide rich semantic structure, but their high ambient dimensionality poses a challenge for NFs.
Let $z \in \mathbb{R}^{N \times n}$ denote the normalized RAE representation in Eq.~\eqref{eq:rae_normalize}, where $N$ is the number of spatial tokens and $n$ is the token channel dimension.

\begin{figure}[t]
    \centering
    \includegraphics[width=\linewidth]{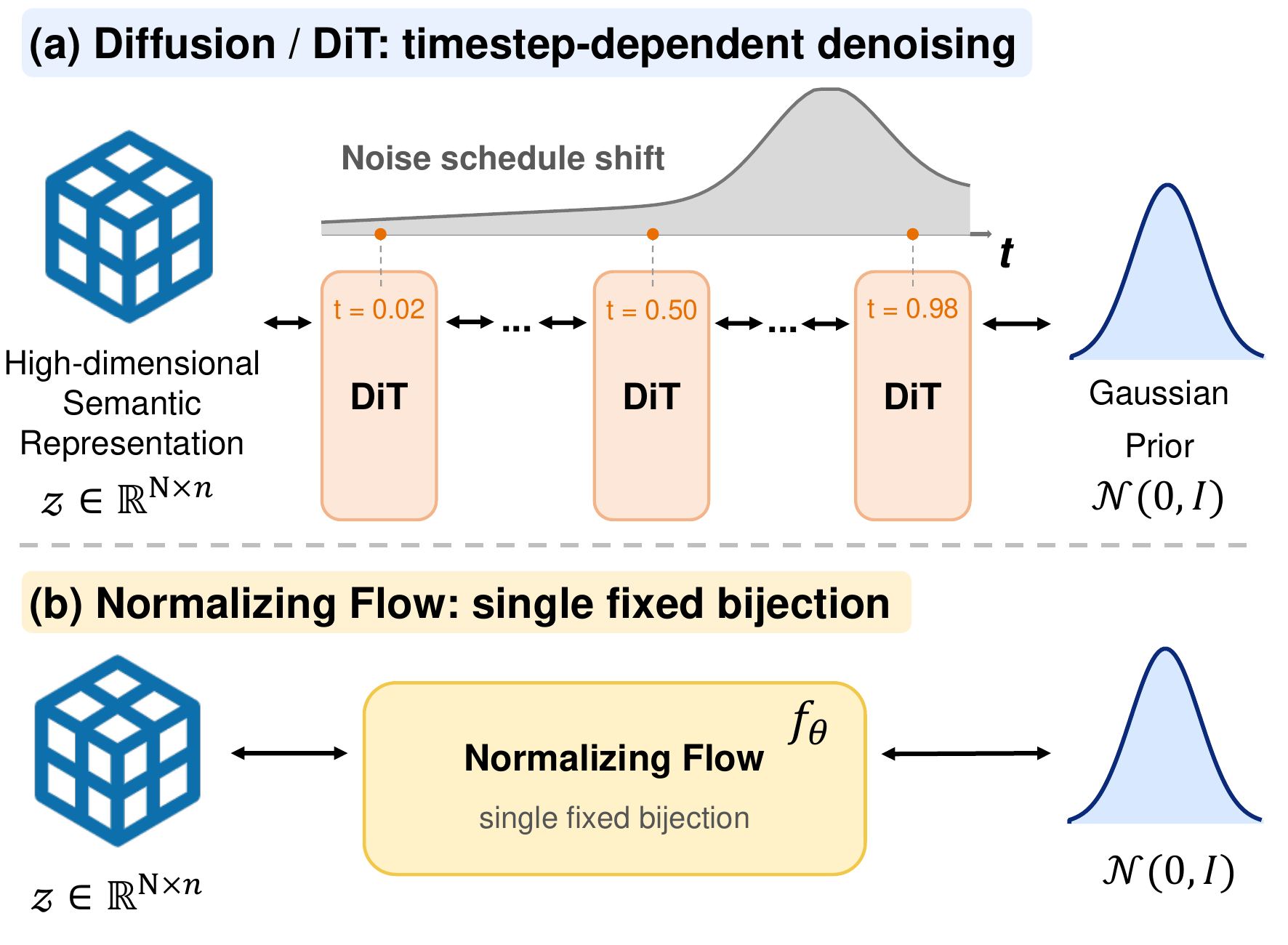}
    \vspace{-2.5em}
    \caption{
    Diffusion adapts through timestep-dependent noise schedule shifts, while NFs learn a single fixed bijection over the full representation space without timestep conditioning.
    }
    \vspace{-1.2em}
    \label{fig:nf_vs_dit}
\end{figure}

Although the effective semantic information is compact, the ambient dimension $Nn$ is large and overcomplete.
For NFs, every modeled channel contributes to the likelihood objective and the log-determinant, forcing the flow to learn an exact invertible transport over the full representation space.
\begin{wraptable}{r}{73.5pt}
    \vspace{-1.0em}
    \hspace*{-14.5pt}\begin{minipage}{88pt}
    \setlength{\tabcolsep}{4pt}
    \centering
    \footnotesize
    \caption{Naive baseline.}
    \vspace{-0.5em}
    \label{tab:naive}
    \begin{tabular}{ccc}
        \toprule
        CFG & Width & gFID$\downarrow$ \\
        \midrule
        \XSolidBrush & 1152 & 11.53 \\
        \XSolidBrush & 2048 & 11.46 \\
        \Checkmark  & 1152 & 3.58 \\
        \Checkmark  & 2048 & 3.54 \\
        \bottomrule
    \end{tabular}
    \end{minipage}
    \vspace{-1.0em}
\end{wraptable}
Diffusion and rectified-flow models, in contrast, learn time-dependent fields and can redistribute learning across noise levels; RAE-based diffusion, for example, calibrates high-dimensional training with a dimension-dependent schedule shift $t^{\prime}=\eta t/(1+(\eta-1)t)$~\citep{zheng2025rae}.
An NF has no such mechanism: it learns a \emph{single fixed bijection} $f_\theta:\mathcal{Y}\to\mathcal{U}$ without a timestep or noise schedule (Figure~\ref{fig:nf_vs_dit}), and must transport the entire distribution of $z\in\mathbb{R}^{N\times n}$ to the Gaussian prior.
This leads to a \emph{semantic-capacity mismatch}: semantic content is compact, but the flow must spend capacity modeling the full ambient representation, including redundant channels.

We empirically verify this mismatch with a naive full-representation baseline.
We train the Transformer autoregressive flow directly on the full normalized RAE tokens.

Since each token requires predicting $2n$ affine parameters for shift and log-scale, we compare the default hidden dimension 1152 with an enlarged 2048.
As shown in Table~\ref{tab:naive}, increasing hidden width brings almost no improvement under either guided or unguided sampling, suggesting that direct full-space modeling is ill-suited for current flow architectures.
The exact compact-space NLL is similarly insensitive to flow width (see Section~\ref{sec:analysis}).
Rather than scaling the flow, we construct a compact semantic representation that preserves the information in RAE features while reducing the ambient dimension seen by the flow.

\subsection{Semantic Representation Compressor}
\label{sec:src}

\paragraph{Compact structure of semantic representations.}
Our analysis shows that directly modeling the full RAE token space is ill-suited for NFs.
Yet pretrained representations are not uniformly informative across channels: they are ambiently high-dimensional but semantically compressible.

\begin{figure}[t]
    \centering
    \includegraphics[width=0.85\linewidth]{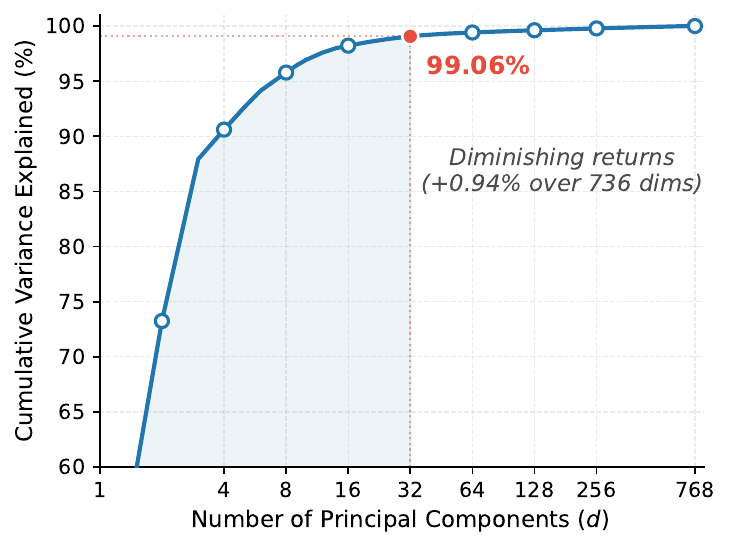}
    \vspace{-1em}
    \caption{
    PCA of normalized RAE features. The first 32 components explain 99.06\% of the total variance.
    }
    \vspace{-1em}
    \label{fig:pca}
\end{figure}

We verify this by performing PCA on normalized RAE features across ImageNet.
As shown in Figure~\ref{fig:pca}, the cumulative explained variance rises rapidly and saturates early: the first 32 principal components already explain 99.06\% of the total variance.
This indicates that most semantic variation can be captured by tens of dimensions rather than the full token dimension $n$.
We therefore seek a learnable compressor that preserves semantic information while reducing the ambient dimensionality modeled by the flow.

\paragraph{SRC architecture.}
We introduce a learnable {Semantic Representation Compressor (SRC)} between the frozen RAE encoder and the normalizing flow.
The SRC contains an encoder $C_{\mathrm{enc}}$ and a decoder $C_{\mathrm{dec}}$.
Given a normalized representation $z \in \mathbb{R}^{N \times n}$, the SRC compresses it to a compact representation and then restores the original dimension:
\begin{equation}
    \tilde{z}_c = C_{\mathrm{enc}}(z) \in \mathbb{R}^{N \times d},
    \qquad
    \hat{z} = C_{\mathrm{dec}}(\tilde{z}_c) \in \mathbb{R}^{N \times n}.
    \label{eq:src_encode_decode}
\end{equation}
The reconstructed representation $\hat{z}$ is denormalized and decoded by the frozen RAE decoder.

A key design choice is to use Transformer blocks for compression.
RAE tokens come from a Vision Transformer encoder and contain long-range semantic correlations across tokens.
Token-wise projections compress each token independently and convolutional compressors capture mainly local interactions; both are less effective in our ablations.

In contrast, Transformer blocks aggregate global token information through self-attention like pretrained vision encoders, which is important because the flow operates only on the compressed representation.

\begin{figure}[t]
    \centering
    \includegraphics[width=\linewidth]{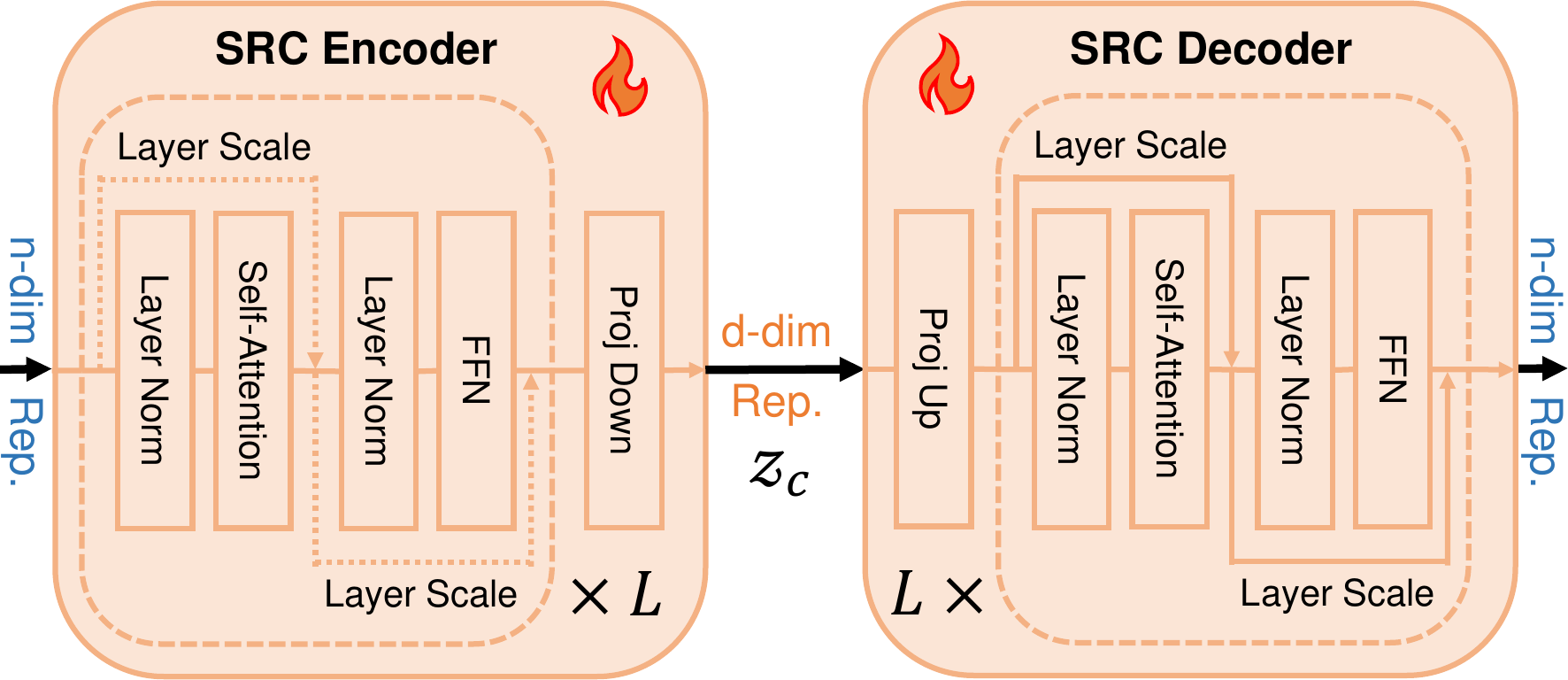}
    \vspace{-1em}
    \caption{
    Semantic Representation Compressor compresses RAE tokens from $n$ to $d$ channels, and restores the original dimension.
    }
    \label{fig:src}
    \vspace{-1em}
\end{figure}

Concretely, the SRC encoder applies $L$ Transformer blocks followed by a projection from $n$ to $d$ channels, and the decoder mirrors it with a projection from $d$ back to $n$ followed by $L$ Transformer blocks (Figure~\ref{fig:src}).

\paragraph{SRC training and compact representation.}
The SRC is trained before the normalizing flow while keeping both the RAE encoder $E$ and decoder $D$ frozen.
Given an image $x$, we obtain $z=\mathrm{norm}(E(x))$, reconstruct it as $\hat{z}=C_{\mathrm{dec}}(C_{\mathrm{enc}}(z))$, and decode $\hat{z}$ through the frozen RAE decoder.
The SRC is optimized with the same reconstruction objective used for RAE decoder training, including pixel, perceptual, and adversarial losses.
Following the RAE protocol, we apply noise augmentation to encoder outputs during SRC training so that the compact representation remains compatible with the noise-tolerant frozen decoder.

After SRC training, we freeze $C_{\mathrm{enc}}$ and $C_{\mathrm{dec}}$.
The normalizing flow is trained on the re-normalized compact semantic representation as follows:
\begin{equation}
    z_c = \mathrm{norm}_2(C_{\mathrm{enc}}(z)),
    \qquad
    z_c \in \mathbb{R}^{N \times d},
    \label{eq:compact_representation}
\end{equation}
where $\mathrm{norm}_2(\cdot)$ denotes channel-wise normalization using statistics computed over compact SRC features.
The compact dimension $d$ controls the trade-off between reconstruction fidelity and flow modeling difficulty.
We use $d=32$ by default, which preserves 99.06\% PCA variance and gives the best full-scale generation performance in our ablation study.

\subsection{Noise Regularization}
\label{sec:noise}

\begin{figure*}[t]
    \centering
    \includegraphics[width=0.90\textwidth]{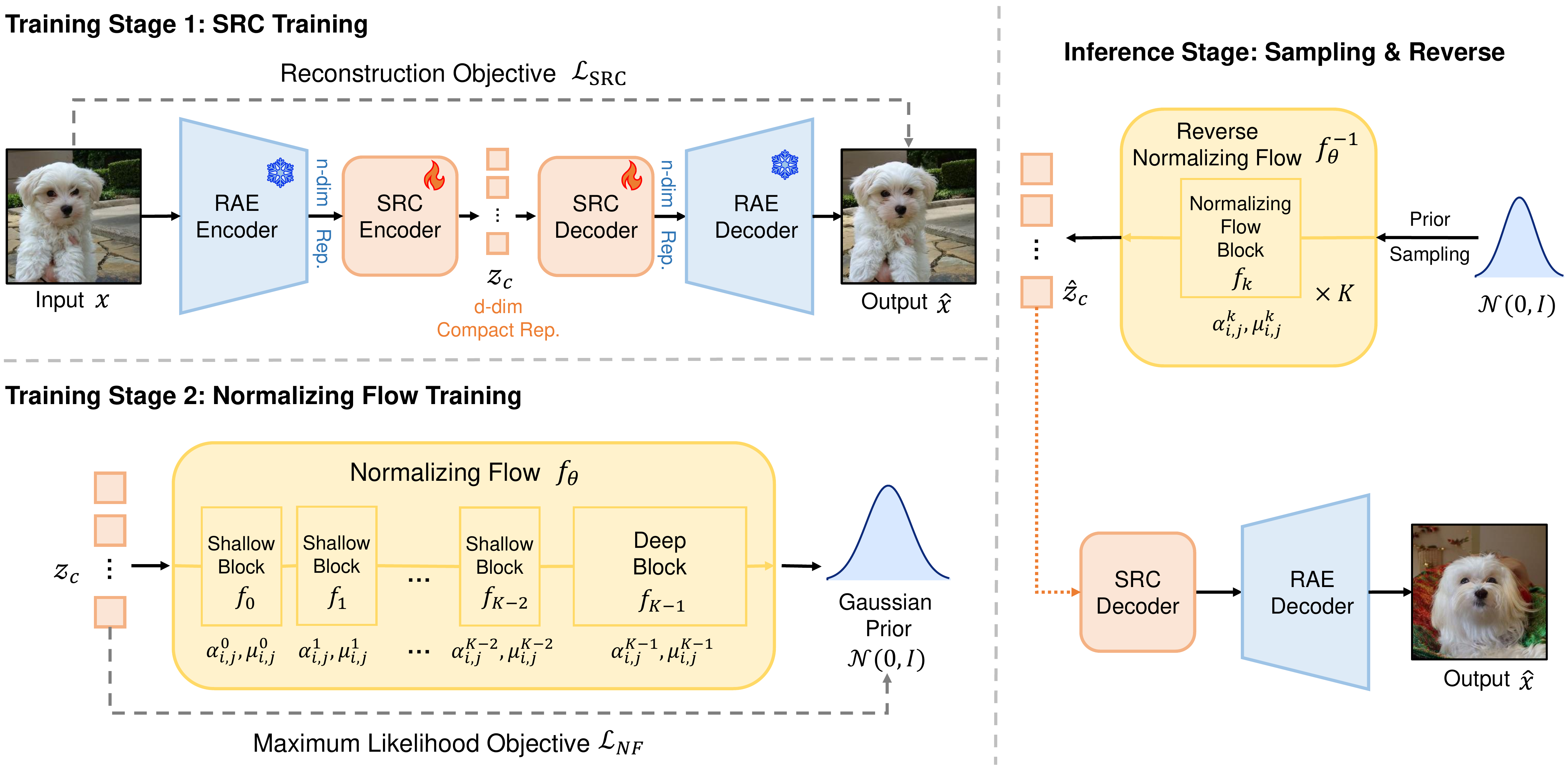}
    \vspace{-0.5em}
    \caption{
    Overview of SRC-Flow.
    Stage 1 trains SRC with frozen RAE.
    Stage 2 trains an NF on compact semantic representations.
    Inference maps Gaussian samples through the inverse NF, the SRC decoder, and the RAE decoder.
    }
    \label{fig:pipeline}
    \vspace{-1.2em}
\end{figure*}

Gaussian noise is a standard regularizer for normalizing flows~\citep{dinh2017density, zhai2024tarflow}: it smooths the empirical data distribution and helps the flow cover the prior more uniformly.
In SRC-Flow, we also inject noise into RAE encoder outputs.

\begin{wraptable}{r}{83.5pt}
    \vspace{-1.0em}
    \centering
    \hspace*{-14.5pt}\begin{minipage}{98pt}
    \footnotesize
    \setlength{\tabcolsep}{2pt}
    \caption{Noise regularization results (gFID$\downarrow$).}

    \label{tab:noise}
    \begin{tabular}{lcc}
        \toprule
        Schedule & w/o cfg & w/ cfg \\
        \midrule
        Per-sample & 11.06 & 1.94 \\
        \rowcolor{gray!15}
        Const.\ ($0.4$) & 8.40 & 1.65 \\
        \bottomrule
    \end{tabular}
    \end{minipage}
    \vspace{-1.0em}
\end{wraptable}
The key design choice is how the noise level is assigned.
The RAE decoder training protocol samples a different perturbation strength for each example, i.e., $\sigma_{\mathrm{flow}}^i \sim \mathcal{U}(0,\sigma_{\mathrm{flow}})$, to improve decoder robustness.
This is suboptimal for NFs: a flow learns a single unconditional bijection that is not conditioned on the perturbation level, so per-sample noise forces it to model a mixture of differently perturbed distributions with one fixed mapping.
We instead use a constant noise level for all samples:
\begin{equation}
    \begin{gathered}
    z_c =
    \mathrm{norm}_2\!\left(C_{\mathrm{enc}}\!\left(\mathrm{norm}(E(x)+\epsilon_{\mathrm{flow}})\right)\right), \\
    \epsilon_{\mathrm{flow}} \sim \mathcal{N}(0,\sigma_{\mathrm{flow}}^2I).
    \end{gathered}
    \label{eq:noisy_compact_representation}
\end{equation}
Constant noise substantially improves both unguided and guided generation (Table~\ref{tab:noise}).
We use $\sigma_{\mathrm{flow}}=0.4$, which performs best under the training schedule (Figure~\ref{fig:noise_sigma}).

\subsection{Overall Pipeline}
\label{sec:pipeline}

SRC-Flow is trained in two stages (Figure~\ref{fig:pipeline}).
We denote the frozen RAE encoder and decoder as $E$ and $D$, the SRC encoder and decoder as $C_{\mathrm{enc}}$ and $C_{\mathrm{dec}}$, and the flow as $f_\theta$.

\paragraph{Stage 1: SRC training.}
With $E$ and $D$ frozen, we train only the SRC using the RAE decoder objective (pixel, perceptual, and adversarial losses) and per-example noise augmentation, following the RAE decoder protocol:
\begin{equation}
    \begin{gathered}
    \hat{x}
    =
    D\!\left(
    \mathrm{denorm}
    \left(
    C_{\mathrm{dec}}
    \left(
    C_{\mathrm{enc}}
    \left(
    \mathrm{norm}(E(x)+\epsilon_{\mathrm{src}})
    \right)
    \right)
    \right)
    \right), \\
    \epsilon_{\mathrm{src}}\sim\mathcal{N}(0,\sigma_{\mathrm{src}}^2 I),\quad
    \sigma_{\mathrm{src}}\sim\mathcal{U}(0,0.8).
    \end{gathered}
    \label{eq:src_pipeline}
\end{equation}
After this stage, both $C_{\mathrm{enc}}$ and $C_{\mathrm{dec}}$ are frozen.

\paragraph{Stage 2: Flow training.}
The flow maps the compact representation $z_c\in\mathbb{R}^{N\times d}$ in Eq.~\eqref{eq:noisy_compact_representation} to a Gaussian prior variable $u=f_\theta(z_c)$ and is trained by maximum likelihood:
\begin{equation}
    \mathcal{L}_{\mathrm{NF}}
    =
    \frac{1}{2}\|f_\theta(z_c)\|_2^2
    +
    \sum_{k=0}^{K-1}
    \sum_{i=0}^{N-1}
    \sum_{j=0}^{d-1}
    \alpha_{i,j}^{k},
    \label{eq:nf_loss}
\end{equation}
where the first term is the negative Gaussian log-density up to a constant and the second is the negative log-determinant accumulated over all flow blocks.

\paragraph{Inference.}
From $u\sim\mathcal{N}(0,I)$, we invert all stages:
\begin{equation}
    \hat{x} = D\!\left(\mathrm{denorm}\left(C_{\mathrm{dec}}\left(\mathrm{denorm}_2\left(f_\theta^{-1}(u)\right)\right)\right)\right).
    \label{eq:inference}
\end{equation}
For class-conditional generation, we use the classifier-free guidance of STARFlow~\citep{gu2025starflow} for autoregressive flows.

\section{Experiments}

\begin{table*}[t]
    \centering
    \caption{Class-conditional image generation on ImageNet $256 \times 256$. SRC-Flow achieves the best result among normalizing flow methods. $^\ddagger$SimFlow continued to 320 epochs with its official configuration (trained by us).}
    \label{tab:main256}
    \footnotesize
    \setlength{\tabcolsep}{8.8pt}
    \begin{tabular}{lcccccccccccc}
        \toprule
        \multirow{2}{*}{Method} & \multirow{2}{*}{Epochs} & \multirow{2}{*}{\#Params} & \multirow{2}{*}{rFID$\downarrow$} & \multicolumn{4}{c}{w/o guidance} & \multicolumn{4}{c}{w/ guidance} \\
        \cmidrule(lr){5-8} \cmidrule(lr){9-12}
        & & & & gFID$\downarrow$ & IS$\uparrow$ & Prec.$\uparrow$ & Rec.$\uparrow$ & gFID$\downarrow$ & IS$\uparrow$ & Prec.$\uparrow$ & Rec.$\uparrow$ \\
        \midrule
        \multicolumn{12}{l}{\emph{Pixel Space}} \\
        \midrule
        ADM~\citep{dhariwal2021diffusion}       & 400  & 554M & -    & 10.94 & 101.0 & 0.69 & 0.63 & 3.94 & 215.8 & 0.83 & 0.53 \\
        PixelFlow~\citep{chen2025pixelflow}     & 320  & 677M & -    & -     & -     & -    & -    & 1.98 & 282.1 & 0.81 & 0.60 \\
        SiD2~\citep{hoogeboom2024sid2}          & 1280 & -    & -    & -     & -     & -    & -    & 1.38 & -     & -    & -    \\
        TARFlow~\citep{zhai2024tarflow}         & 320  & 1.4B & -    & -     & -     & -    & -    & 4.69 & -     & -    & -    \\
        iTARFlow~\citep{chen2026itarflow}       & 320  & 2.2B & -    & -     & -     & -    & -    & 3.32 & -     & -    & -    \\
        JetFormer~\citep{tschannen2024jetformer}& 500  & 2.8B & -    & -     & -     & -    & -    & 6.64 & -     & 0.69 & 0.56 \\
        FARMER~\citep{zheng2025farmer}          & 320  & 1.9B & -    & -     & -     & -    & -    & 3.60 & 269.2 & 0.81 & 0.51 \\
        \midrule
        \multicolumn{12}{l}{\emph{Latent Autoregressive}} \\
        \midrule
        VAR~\citep{tian2024var}                 & 350  & 2.0B & -    & 1.92  & 323.1 & 0.82 & 0.59 & 1.73 & 350.2 & 0.82 & 0.60 \\
        MAR~\citep{li2024mar}                   & 800  & 943M & 0.53 & 2.35  & 227.8 & 0.79 & 0.62 & 1.55 & 303.7 & 0.81 & 0.62 \\
        xAR~\citep{ren2025xar}                  & 800  & 1.1B & 0.53 & -     & -     & -    & -    & 1.24 & 301.6 & 0.83 & 0.64 \\
        \midrule
        \multicolumn{12}{l}{\emph{Latent Diffusion}} \\
        \midrule
        DiT~\citep{peebles2023scalable}         & 1400 & 675M & 0.61 & 9.62  & 121.5 & 0.67 & 0.67 & 2.27 & 278.2 & 0.83 & 0.57 \\
        SiT~\citep{ma2024sit}                   & 1400 & 675M & 0.61 & 8.61  & 131.7 & 0.68 & 0.67 & 2.06 & 270.3 & 0.82 & 0.59 \\
        REPA~\citep{yu2024repa}                 & 800  & 675M & 0.61 & 5.78  & 158.3 & 0.70 & 0.68 & 1.29 & 306.3 & 0.79 & 0.64 \\
        VA-VAE~\citep{yao2025vavae}             & 800  & 675M & 0.28 & 2.17  & 205.6 & 0.77 & 0.65 & 1.35 & 295.3 & 0.79 & 0.65 \\
        DDT~\citep{wang2025ddt}                 & 400  & 675M & 0.61 & 6.27  & 154.7 & 0.68 & 0.69 & 1.26 & 310.6 & 0.79 & 0.65 \\
        REPA-E~\citep{leng2025repae}            & 800  & 675M & 0.28 & 1.69  & 219.3 & 0.77 & 0.67 & 1.12 & 302.9 & 0.79 & 0.66 \\
        RAE~\citep{zheng2025rae}                & 800  & 839M & 0.57 & 1.51  & 242.9 & 0.79 & 0.63 & 1.13 & 262.6 & 0.78 & 0.67 \\
        \midrule
        \multicolumn{12}{l}{\emph{Latent Normalizing Flows}} \\
        \midrule
        FlowBack~\citep{chen2025flowing}         & 320 & 831M & 0.61 & -     & -     & -    & -    & 4.18 & 240.8 & -    & -    \\
        STARFlow~\citep{gu2025starflow}         & 320 & 1.4B & 2.73 & -     & -     & -    & -    & 2.40 & -     & -    & -    \\
        iTARFlow~\citep{chen2026itarflow}       & 320 & 770M & 2.73 & -     & -     & -    & -    & 2.32 & -     & -    & -    \\
        SimFlow~\citep{zhao2025simflow}         & 160 & 1.4B & 1.08 & 10.13 & 124.7 & 0.71 & 0.61 & 1.91 & 284.4 & 0.82 & 0.60 \\
        SimFlow$^\ddagger$                      & 320 & 1.4B & 1.08 & 9.37  & 125.6 & -    & -    & 1.90 & 285.2 & 0.82 & 0.60 \\
        \rowcolor{gray!15}
        SRC-Flow (Ours)                        & 160 & 1.4B & 0.62 & 9.28  & 125.1 & -    & -    & 1.73 & 288.9 & 0.82 & 0.61 \\
        \rowcolor{gray!15}
        SRC-Flow (Ours)                        & 320 & 1.4B & 0.62 & {8.40} & {127.4} & {0.72} & {0.61} & {1.65} & {289.7} & {0.82} & {0.61} \\
        \bottomrule
    \end{tabular}
    \vspace{-1em}
\end{table*}

\subsection{Experimental Setup}
\label{sec:setup}

\paragraph{Dataset, metrics, and baselines.}
We evaluate class-conditional generation on ImageNet~\citep{deng2009imagenet} at $256\times256$ and $512\times512$.
Generation quality is measured by gFID~\citep{heusel2017gans}, Inception Score (IS)~\citep{salimans2016improved}, Precision, and Recall~\citep{kynkaanniemi2019improved}, using 50k samples and ADM statistics~\citep{dhariwal2021diffusion}.
Reconstruction is measured by rFID on ImageNet validation.
We compare with pixel-space, latent autoregressive, latent diffusion, and latent NF models.
Among NFs, TARFlow~\citep{zhai2024tarflow} operates in pixel space; FlowBack~\citep{chen2025flowing} aligns reverse-pass features in a latent TARFlow variant; STARFlow~\citep{gu2025starflow}, iTARFlow~\citep{chen2026itarflow}, and SimFlow~\citep{zhao2025simflow} improve latent flow modeling via architectural, denoising, or training modifications.

\paragraph{Implementation details.}
The SRC uses compact dimension $d=32$ and $L=4$ Transformer layers.
We train SRC for 16 epochs with batch size 256 using AdamW~\citep{loshchilov2019adamw}, learning rate $2\times10^{-4}$, cosine decay after 1 warm epoch.
The flow follows the SimFlow~\citep{zhao2025simflow} architecture and is trained for 320 epochs with batch size 256, AdamW, learning rate $1\times10^{-4}$, cosine decay from epoch 160, EMA decay 0.9999, and a constant noise level $\sigma_{\mathrm{flow}}=0.4$.
Class labels are dropped with probability 0.1 during training, and we sample with a classifier-free guidance weight $\omega=1.2$.

\subsection{Main Results}
\label{sec:main_results}

\begin{figure*}[t]
    \centering
    \includegraphics[width=0.91\textwidth]{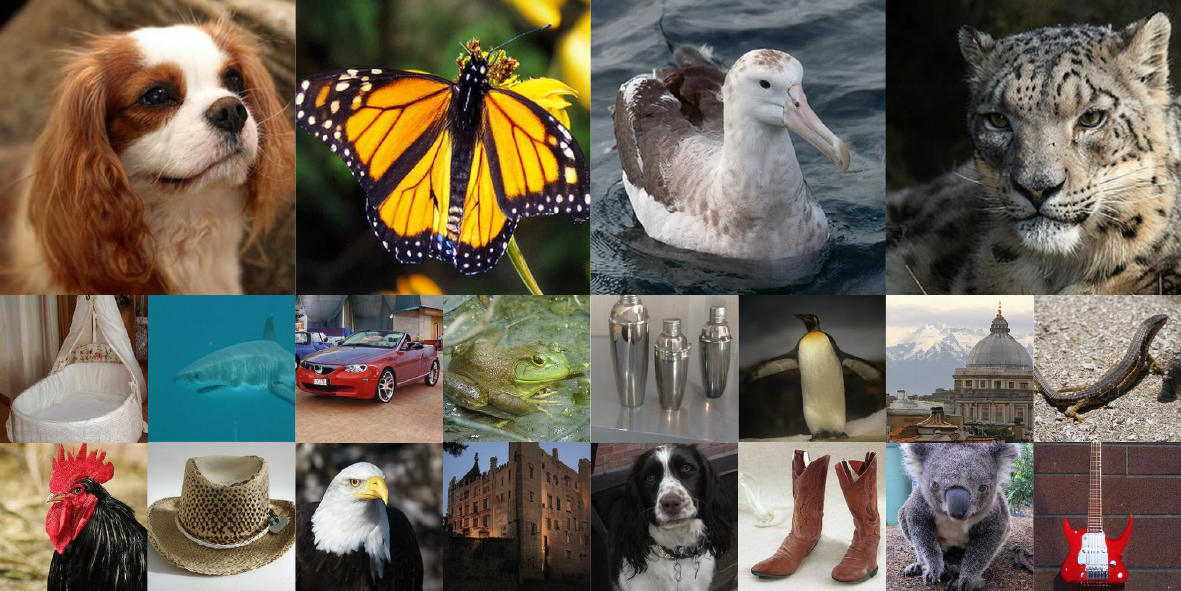}
    \caption{Class-conditional SRC-Flow samples on ImageNet: $512\times512$ (top row) and $256\times256$ (other rows).
    }
    \label{fig:samples}
    \vspace{-1em}
\end{figure*}

\paragraph{Results on ImageNet $256 \times 256$.}
Among normalizing flows, SRC-Flow achieves the best guided gFID of 1.65 (Table~\ref{tab:main256}), outperforming pixel-space NFs such as TARFlow (4.69), FARMER (3.60), and iTARFlow (3.32), and latent-space NFs such as FlowBack (4.18), STARFlow (2.40), latent iTARFlow (2.32), and SimFlow (1.91); without guidance, it improves the best NF result from 10.13 (SimFlow) to 8.40.
Since SimFlow reports 160 epochs, we additionally continue training SimFlow to 320 epochs with its official configuration and report SRC-Flow at 160 epochs.

\begin{table}[t]
    \centering
    \footnotesize
    \setlength{\tabcolsep}{12pt}
    \caption{ImageNet $512\times512$ w/ guidance. SRC-Flow achieves best result among normalizing flow methods.}
    \label{tab:main512}
    \renewcommand{\arraystretch}{1.1}
    \begin{tabular}{lccc}
        \toprule
        Method & \#Params & gFID$\downarrow$ & IS$\uparrow$ \\
        \midrule
        BigGAN-deep~\citep{brock2019biggan}     & 158M & 8.43 & 177.9 \\
        StyleGAN-XL~\citep{sauer2022styleganxl} & -    & 2.41 & 267.8 \\
        \midrule
        VAR~\citep{tian2024var}                 & 2.3B & 2.63 & 303.2 \\
        MAGVIT-v2~\citep{yu2024magvitv2}        & 307M & 1.91 & 324.3 \\
        MAR~\citep{li2024mar}                   & 481M & 1.73 & 279.9 \\
        xAR~\citep{ren2025xar}                  & 608M & 1.70 & 281.5 \\
        \midrule
        ADM~\citep{dhariwal2021diffusion}       & 731M & 3.85 & 221.7 \\
        SiD2~\citep{hoogeboom2024sid2}          & -    & 1.50 & - \\
        DiT~\citep{peebles2023scalable}         & 674M & 3.04 & 240.8 \\
        SiT~\citep{ma2024sit}                   & 674M & 2.62 & 252.2 \\
        DiffiT~\citep{hatamizadeh2024diffit}    & -    & 2.67 & 252.1 \\
        REPA~\citep{yu2024repa}                 & 675M & 2.08 & 274.6 \\
        DDT~\citep{wang2025ddt}                 & 675M & 1.28 & 305.1 \\
        EDM2~\citep{karras2024edm2}             & 1.5B & 1.25 & - \\
        RAE~\citep{zheng2025rae}                & 839M & 1.13 & 259.6 \\
        \midrule
        STARFlow~\citep{gu2025starflow}         & 1.4B & 3.00 & - \\
        SimFlow~\citep{zhao2025simflow}         & 1.4B & 2.74 & 304.9 \\
        \rowcolor{gray!15}
        SRC-Flow                               & 1.4B & {2.07} & {305.7} \\
        \bottomrule
    \end{tabular}
    \vspace{-3em}
\end{table}

At matched budgets, SRC-Flow improves guided gFID from 1.91 to 1.73 at 160 epochs and from 1.90 to 1.65 at 320 epochs, whereas longer training leaves SimFlow essentially unchanged, and the gFID ranges over 10 independent 50k evaluation sets do not overlap between the two methods.

Although SRC reduces the modeled channel dimension from $n$ to 32, its rFID of 0.62 remains close to the 0.57 of the original RAE, indicating that SRC preserves decoder-relevant semantic information while substantially reducing the ambient dimensionality faced by the flow.

\paragraph{Results on ImageNet $512 \times 512$.}

SRC-Flow achieves gFID 2.07 and IS 305.7 (Table~\ref{tab:main512}), clearly outperforming STARFlow (3.00) and SimFlow (2.74), while SRC reconstruction stays close to the RAE (rFID 0.54 vs.\ 0.53).

\paragraph{Qualitative samples on ImageNet $256 \times 256$ and $512 \times 512$.}
Figure~\ref{fig:samples} shows class-conditional SRC-Flow samples at both resolutions with classifier-free guidance.

\subsection{Analysis}
\label{sec:analysis}

\begin{figure*}[t]
    \centering
    \includegraphics[width=\textwidth]{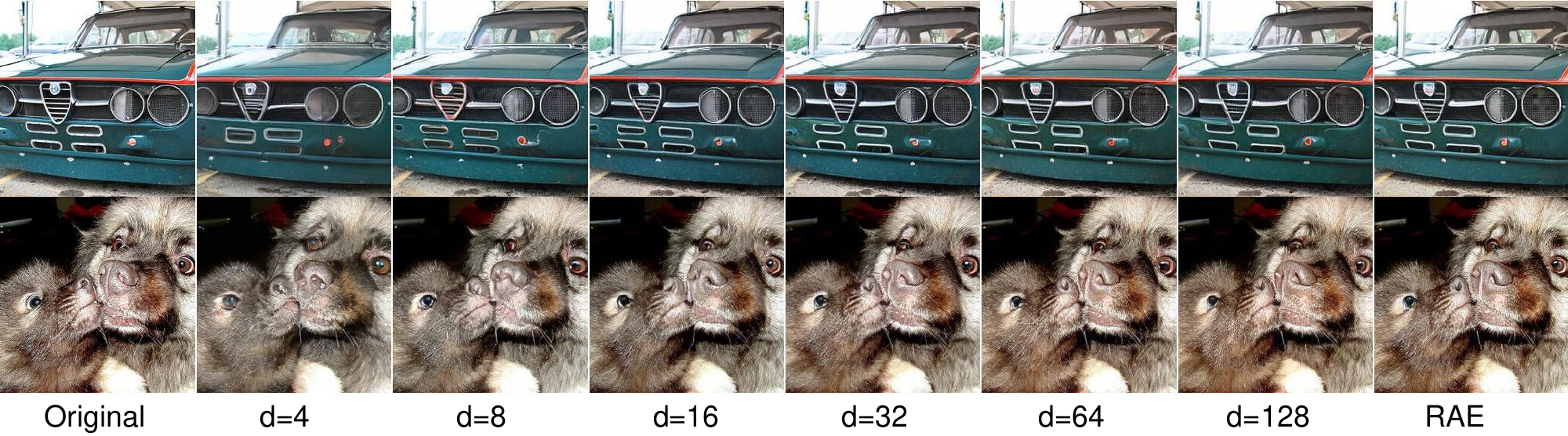}
    \vspace{-2em}
    \caption{
    Reconstructions across $d$. The $d=32$ SRC preserves visual details close to the original RAE.
    }
    \label{fig:dim_recon_vis}
    \vspace{-1em}
\end{figure*}

\paragraph{Compact representation dimension.}
We first study the compact dimension $d$.
As shown in Table~\ref{tab:dim}, reconstruction fidelity improves as $d$ increases, since larger compact representations preserve more decoder-relevant information.
Semantic preservation follows the same trend: at $d=32$, Top-5 accuracy of reconstructions remains 89.8\% (vs.\ 96.6\% for the full RAE), although Top-1 continues to improve with larger $d$, so not every fine-grained cue is preserved; DINO and CLIP similarities show the same saturation as $d$ grows.
However, generation quality is not monotonic: small $d$ loses useful semantic and visual information, while large $d$ increases the ambient dimensionality that the flow must model under exact likelihood.
The best balance is at $d=32$ (gFID 1.65, rFID 0.62 vs.\ 0.57 for the RAE), supporting our claim that the full RAE space is unnecessarily overcomplete for flow modeling; Figure~\ref{fig:dim_recon_vis} shows that its reconstructions remain visually close to those of the full RAE.

\begin{table}[t]
    \centering
    \footnotesize
    \setlength{\tabcolsep}{3pt}
    \caption{Effect of compact dimension $d$ on ImageNet $256\times256$. Top-1/Top-5 is the accuracy of a separately trained, frozen MLP classifier on reconstructed images, measuring semantic preservation. $d=768$ denotes the full RAE representation.}
    \vspace{-1em}
    \label{tab:dim}
    \begin{tabular}{lccccc}
        \toprule
        & \multicolumn{2}{c}{Reconstruction \& semantics} & \multicolumn{2}{c}{gFID$\downarrow$} \\
        \cmidrule(lr){2-3} \cmidrule(lr){4-5}
        $d$ & rFID$\downarrow$ & Top-1/Top-5$\uparrow$ & w/o cfg & w/ cfg \\
        \midrule
        4   & 1.13 & 7.3/19.6  & 15.94 & 3.53 \\
        8   & 0.75 & 20.7/43.2 & 7.23  & 2.12 \\
        16  & 0.64 & 47.3/70.4 & 7.84  & 1.69 \\
        \rowcolor{gray!15}
        32  & 0.62 & 69.1/89.8 & 8.40  & 1.65 \\
        64  & 0.60 & 74.7/93.8 & 9.78  & 1.97 \\
        128 & 0.60 & 78.9/95.1 & 10.65 & 2.21 \\
        768 (RAE) & 0.57 & 82.6/96.6 & 11.53 & 3.58 \\
        \bottomrule
    \end{tabular}
    \vspace{-1em}
\end{table}

\begin{table}[t]
    \centering
    \footnotesize
    \setlength{\tabcolsep}{2.5pt}
    \begin{minipage}[t]{0.505\linewidth}
    \centering
    \setlength{\tabcolsep}{2.5pt}
    \renewcommand{\arraystretch}{1.25}
    \captionof{table}{Compressor architecture (all at $d=32$). Stronger token interaction performs better.}
    \vspace{-1em}
    \label{tab:compressor_arch}
    \begin{tabular}{lccc}
        \toprule
        Comp. & rFID$\downarrow$ & gFID$\downarrow$ & IS$\uparrow$ \\
        \midrule
        PCA        & 1.08 & 3.31 & 263.8 \\
        Linear     & 0.94 & 2.86 & 274.5 \\
        Conv       & 0.70 & 2.14 & 282.8 \\
        \rowcolor{gray!15}
        Trans (SRC) & 0.62 & 1.65 & 289.7 \\
        \bottomrule
    \end{tabular}
    \end{minipage}\hfill
    \begin{minipage}[t]{0.465\linewidth}
    \centering
    \setlength{\tabcolsep}{4pt}
    \renewcommand{\arraystretch}{1.05}
    \captionof{table}{SRC depth. Quality improves up to $L=4$; deeper ones barely improve rFID.}
    \vspace{-1em}
    \label{tab:compressor_depth}
    \begin{tabular}{lccc}
        \toprule
        $L$ & rFID$\downarrow$ & gFID$\downarrow$ & IS$\uparrow$ \\
        \midrule
        1  & 0.77 & 2.34 & 282.6 \\
        2  & 0.68 & 1.87 & 291.4 \\
        \rowcolor{gray!15}
        4  & 0.62 & 1.65 & 289.7 \\
        8  & 0.60 & 1.65 & 289.1 \\
        16 & 0.59 & 1.66 & 289.6 \\
        \bottomrule
    \end{tabular}
    \end{minipage}
    \vspace{-2em}
\end{table}

\begin{figure}[t]
    \centering
    \includegraphics[width=\linewidth]{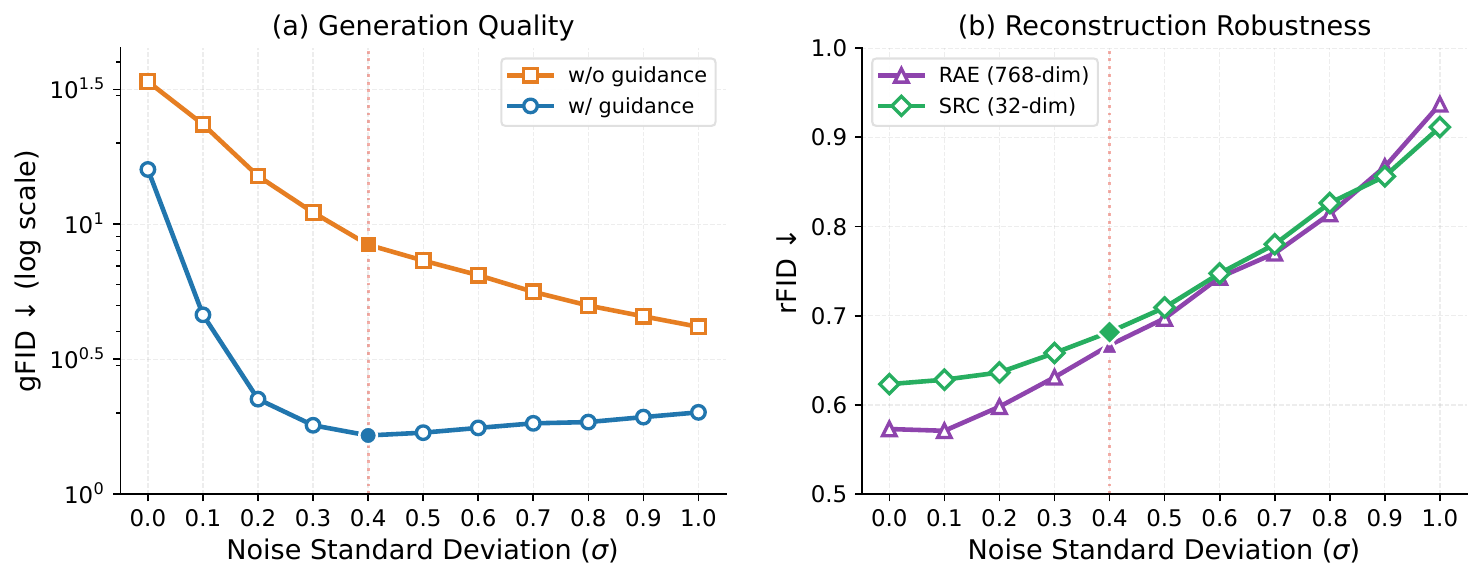}
    \vspace{-2em}
    \caption{
    Noise regularization.
    $\sigma_{\mathrm{flow}}=0.4$ gives the best gFID, and the $d=32$ SRC improves high-noise robustness.
    }
    \label{fig:noise_sigma}
    \vspace{-1em}
\end{figure}

\paragraph{Compressor architecture and depth.}
Table~\ref{tab:compressor_arch} compares compression designs.
Non-learned PCA already improves on the full RAE (3.31 vs.\ 3.58) without training a new tokenizer, but is not optimized for the frozen decoder or the flow.
Learnable compressors perform better, and stronger token interaction helps further: token-wise linear projection is weaker than local convolution, and the Transformer-based SRC performs best by aggregating information across tokens with global self-attention.
Semantic compression thus benefits from modeling global token dependencies rather than mere dimensionality reduction.
A noise-robust VA-VAE trained with the same NF and schedule reaches gFID 1.89, still behind SRC.
For depth (Table~\ref{tab:compressor_depth}), increasing $L$ from 1 to 4 improves reconstruction and generation, while deeper compressors only slightly improve rFID with negligible generation gains; we use $L=4$ by default.

\paragraph{Noise regularization.}
Training the flow without noise converges poorly (Figure~\ref{fig:noise_sigma}), confirming that noise regularization is essential in compact semantic space, and $\sigma_{\mathrm{flow}}=0.4$ performs best.
In terms of reconstruction robustness, the $d=32$ SRC stays close to the full RAE at low noise and even achieves lower rFID under strong perturbations, suggesting that compression suppresses redundant, noise-sensitive directions and acts as an implicit denoising projection.

\paragraph{Transfer across pretrained encoders.}
To test whether the compact regime is specific to DINOv2, we train SRC and the same flow on the released SigLIP2-~\citep{tschannen2025siglip} and MAE-based~\citep{he2022mae} RAEs~\citep{zheng2025rae}, with a matched 160-epoch schedule and $\sigma_{\mathrm{flow}}=0.4$ without retuning (Table~\ref{tab:encoders}).
All three encoders show a compact optimum or saturation near $d=32$: increasing to $d=64$ improves rFID but not generation.
Notably, MAE reconstructs best yet generates worst, again showing that reconstruction fidelity alone does not determine NF tractability.
Absolute quality and the precise optimum remain encoder-dependent, so we do not claim that $d=32$ is universal across all pretrained encoders.

\paragraph{Compact-space likelihood.}
Exact NLL in the compact space mirrors the generation trends.
Widening the full-RAE flow from 1152 to 2048 changes NLL only from 0.850 to 0.838 nats/dim, consistent with its unchanged gFID, whereas compressing to $d=32$ lowers NLL to 0.409.
NLL is lowest at $d\in\{8,16\}$, yet generation peaks at $d=32$: overly compact variables are easy to fit but discard semantics, so NLL diagnoses flow tractability rather than replacing sample-quality metrics such as gFID and IS.

\begin{table}[t]
    \centering
    \footnotesize
    \setlength{\tabcolsep}{15pt}
    \caption{SRC-Flow with different RAE encoders at $d=16/32/64$ (160 epochs, $\sigma_{\mathrm{flow}}=0.4$); best gFID in bold.}
    \vspace{-1em}
    \label{tab:encoders}
    \begin{tabular}{lcc}
        \toprule
        Encoder & rFID$\downarrow$ & w/cfg gFID$\downarrow$ \\
        \midrule
        DINOv2  & 0.64 / 0.62 / 0.60 & 1.82 / \textbf{1.73} / 2.01 \\
        SigLIP2 & 0.97 / 0.91 / 0.87 & 2.12 / \textbf{2.05} / 2.32 \\
        MAE     & 0.58 / 0.43 / 0.36 & 3.72 / \textbf{3.36} / 3.37 \\
        \bottomrule
    \end{tabular}
    \vspace{-1em}
\end{table}

\paragraph{Training and sampling cost.}
SRC training adds only 10.5 hours (4.0\% of the 320-epoch budget), and from released RAE checkpoints SRC-Flow is cheaper than SimFlow at both budgets.
Sampling is slow at batch size 1 (0.03 images/s) but reaches 1.32 images/s at batch 128, above the evaluated autoregressive and diffusion baselines.

\section{Conclusion}
\label{sec:conclusion}
We presented SRC-Flow, which identifies a semantic-capacity mismatch between the compact semantics of pretrained representations and the full ambient space that NFs must model.
Our SRC, together with constant noise regularization, reduces this modeling burden and achieves state-of-the-art quality among NF methods, with exact likelihood in the compact space and deterministic invertible sampling.
\textbf{Limitations and future work.}
Exact likelihood holds for the compact representation rather than for images, since the decoders are not invertible.
Our experiments are class-conditional on ImageNet, a gap remains to the strongest diffusion and autoregressive models, autoregressive sampling limits throughput, and the frozen RAE decoder bounds reconstruction fidelity.
Future work includes non-autoregressive flows, stronger compression objectives, and text-conditional or higher-resolution image generation.

{
    \small
    \bibliographystyle{ieeenat_fullname}
    \bibliography{ref}
}

\end{document}

%% file: preamble.tex
\usepackage[utf8]{inputenc}
\usepackage[T1]{fontenc}
\usepackage{amsfonts}
\usepackage{microtype}
\usepackage{colortbl}
\usepackage{multirow}
\usepackage{capt-of}
\usepackage{wrapfig}
\IfFileExists{bbding.sty}{\usepackage{bbding}}{\newcommand{\XSolidBrush}{$\times$}\newcommand{\Checkmark}{$\checkmark$}}